\documentclass[letterpaper]{article} 
\usepackage{aaai23}  
\usepackage{times}  
\usepackage{helvet}  
\usepackage{courier}  
\usepackage[hyphens]{url}  
\usepackage{graphicx} 
\usepackage{natbib}  
\usepackage{caption} 
\usepackage{algorithm}
\usepackage{algorithmic}

\usepackage{newfloat}
\usepackage{listings}
\DeclareCaptionStyle{ruled}{labelfont=normalfont,labelsep=colon,strut=off} 
\floatstyle{ruled}
\newfloat{listing}{tb}{lst}{}
\floatname{listing}{Listing}

\usepackage{amsmath}
\usepackage{amssymb}
\usepackage{booktabs}
\usepackage{multirow}
\usepackage{array}
\usepackage{makecell}

\title{KPT: Keyword-guided Pre-training for Grounded Dialog Generation}
\author{
    Qi Zhu\textsuperscript{\rm 1}\thanks{Work done during an internship at Huawei Noah's Ark Lab.},
    Fei Mi\textsuperscript{\rm 2}\footnotemark[2],
    Zheng Zhang\textsuperscript{\rm 1},
    Yasheng Wang\textsuperscript{\rm 2},
    Yitong Li\textsuperscript{\rm 2},\\
    Xin Jiang\textsuperscript{\rm 2},
    Qun Liu\textsuperscript{\rm 2},
    Xiaoyan Zhu\textsuperscript{\rm 1},
    Minlie Huang\textsuperscript{\rm 1}\footnote{Corresponding authors.}
}
\affiliations{
    \textsuperscript{\rm 1}CoAI Group, DCST, IAI, BNRIST, Tsinghua University\\
    \textsuperscript{\rm 2}Huawei Noah’s Ark Lab
    
    zhu-q18@mails.tsinghua.edu.cn,
    \{z-zhang,zxy-dcs,aihuang\}@tsinghua.edu.cn
    \{mifei2,wangyasheng,liyitong3,Jiang.Xin,qun.liu\}@huawei.com
}

\usepackage{bibentry}

\begin{document}

\maketitle

\begin{abstract}
Incorporating external knowledge into the response generation process is essential to building more helpful and reliable dialog agents.
However, collecting knowledge-grounded conversations is often costly, calling for a better pre-trained model for grounded dialog generation that generalizes well w.r.t. different types of knowledge.
In this work, we propose \textbf{KPT} (\textbf{K}eyword-guided \textbf{P}re-\textbf{T}raining), a novel self-supervised pre-training method for grounded dialog generation without relying on extra knowledge annotation.
Specifically, we use a pre-trained language model to extract the most uncertain tokens in the dialog as \emph{keywords}.
With these keywords, we construct two kinds of knowledge and pre-train a knowledge-grounded response generation model, aiming at handling two different scenarios: (1) the knowledge should be faithfully grounded; (2) it can be selectively used.
For the former, the grounding knowledge consists of keywords extracted from the response.
For the latter, the grounding knowledge is additionally augmented with keywords extracted from other utterances in the same dialog.
Since the knowledge is extracted from the dialog itself, KPT can be easily performed on a large volume and variety of dialogue data. 
We considered three data sources (open-domain, task-oriented, conversational QA) with a total of 2.5M dialogues.
We conduct extensive experiments on various few-shot knowledge-grounded generation tasks, including grounding on dialog acts, knowledge graphs, persona descriptions, and Wikipedia passages.
Our comprehensive experiments and analyses demonstrate that KPT consistently outperforms state-of-the-art methods on these tasks with diverse grounding knowledge.
\end{abstract}

\section{Introduction}

\begin{figure}[t]
\centering
\includegraphics[width=0.8\columnwidth]{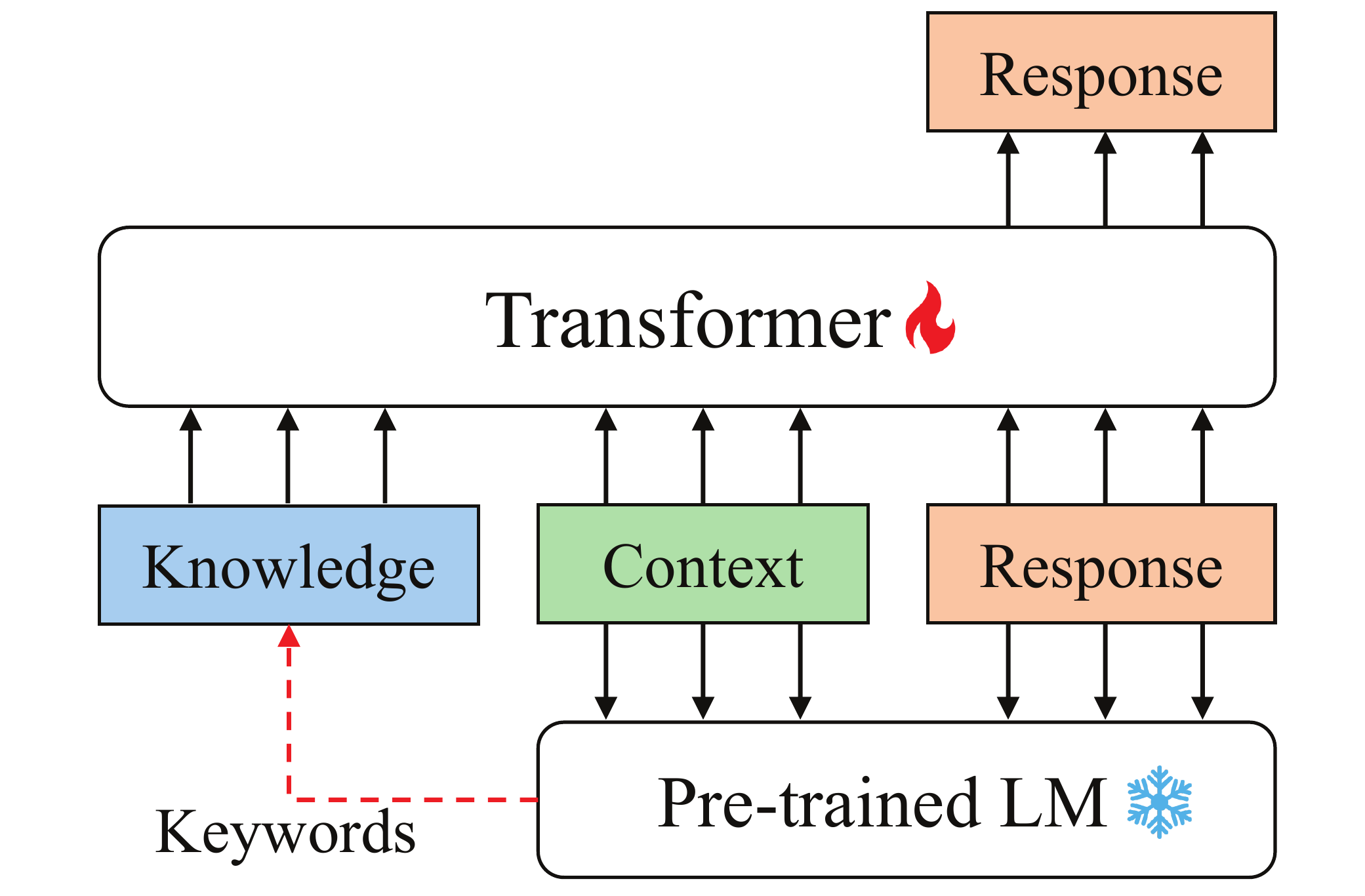} 
\caption{An illustration of \textit{keyword-guided pre-training}. We extract the most uncertain tokens in a dialog (measured by the bottom frozen pre-trained LM) as keywords and create pseudo-knowledge from these keywords. The pre-training objective (upper) is to generate the response according to the context and knowledge.}
\label{fig:illustration}
\end{figure}

Grounding the response generation on external knowledge is a necessary skill of an engaging and reliable dialog agent \cite{shuster-etal-2021-retrieval-augmentation,thoppilan2022lamda,shuster2022blenderbot}.
Such a dialog agent could handle various scenarios by grounding on relevant knowledge such as common sense knowledge graphs \cite{zhou2018commonsense}, persona descriptions \cite{zhang-etal-2018-personalizing},  and Wikipedia passages \cite{dinan2018wizard}.
However, the collection of human-annotated knowledge-grounded dialogs is expensive, which highlights the model's few-shot learning ability to ground on new kinds of knowledge.
To this end, we propose \textbf{K}eyword-guided \textbf{P}re-\textbf{T}raining (KPT), a self-supervised pre-training method that only needs dialogs without any extra knowledge annotation but benefits multiple downstream dialog generation tasks grounded on different knowledge sources.

Most of the previous dialog pre-training methods aim to capture the coherence of dialog, such as predicting the next utterance \cite{zhang-etal-2020-dialogpt}, masked utterance \cite{mehri-etal-2019-pretraining}, and correct utterance order \cite{gu2021dialogbert}.
However, for grounded dialog generation, learning to incorporate external knowledge for a given dialog context is more critical.
Recently proposed GODEL \cite{peng2022godel} utilizes several annotated grounded dialog corpora for pre-training by concatenating the knowledge and context as input.
In contrast, our proposed KPT can be performed on any dialog corpus without requiring grounding annotation, which could benefit more downstream tasks by including more diverse dialogs in pre-training.

Pre-training a general knowledge-grounded dialog generation model faces two main challenges: the diverse knowledge types of downstream tasks and the lack of large annotated knowledge-grounded dialog datasets.
To address these challenges, we regard the keywords as the basic unit of knowledge and mine the keywords from the dialog itself as \citeauthor{feng-etal-2021-language} \shortcite{feng-etal-2021-language}.
As shown in Figure \ref{fig:illustration}, we extract the most uncertain tokens measured by a pre-trained language model (LM) as keywords and construct pseudo-knowledge input using these keywords. 
We identify two scenarios of grounded dialog generation and design two pre-training tasks accordingly which only differ in knowledge construction:
(1) models need to faithfully ground on the given knowledge, where we use keywords from the response as the knowledge; (2) models need to selectively use part of (or even none of) the knowledge based on the context, where we use keywords sampled from other turns of the same dialog along with (or not) keywords from the response as the knowledge.
We pre-train the models jointly on these two tasks on 2.5M dialogs covering open-domain dialogs, task-oriented dialogs, and conversational question answering.
To verify the effectiveness of KPT, we conduct few-shot fine-tuning experiments on four grounded dialog generation tasks where the knowledge sources include dialog acts, knowledge graphs, persona descriptions, and Wikipedia passages.
We show that KPT consistently outperforms GODEL and standard response generation pre-training.
We further analyze the effects of each pre-training task, keyword ratio, and the pre-trained LM used for keyword extraction.

In summary, our contributions are:
\begin{enumerate}
    \item We propose KPT, a novel self-supervised dialog pre-training method for grounded dialog generation, which does not need any extra knowledge annotation.
    \item We identify two scenarios of grounded dialog generation and design corresponding pre-training objectives to handle them in a unified framework.
    \item Experiments on diverse grounded dialog generation tasks with different knowledge sources show that our proposed KPT consistently outperforms state-of-the-art baselines.
\end{enumerate}









\section{Related Work}
\subsection{Pre-training}
Pre-training is a widely used technique to transfer knowledge from a source domain to a target domain and is especially effective when the target domain data is limited.
In recent years, big models pre-trained on large-scale text corpora have been applied to various downstream tasks and achieved remarkable success \cite{devlin-etal-2019-bert,radford2019gpt2,raffel2020t5,lewis-etal-2020-bart}.
To further improve the performance on dialog modeling, researchers perform pre-training on dialog corpora to reduce domain gap \cite{zhang-etal-2020-dialogpt,DeFreitas2020meena,roller-etal-2021-recipes,wu-etal-2020-tod,PanGuBot-feimi-2022} and devise novel pre-training objectives to capture dialog-specific features \cite{mehri-etal-2019-pretraining,gu2021dialogbert,xu-zhao-2021-dialogue}.
However, most proposed self-supervised pre-training objectives focus on learning the internal characteristics of dialog data instead of how to incorporate external knowledge into the dialog.

To learn a knowledge-grounded dialog generation model, existing works either use human-annotated knowledge-grounded conversation corpora for training \cite{thoppilan2022lamda,peng-etal-2021-soloist,peng2022godel,he2022unified} or augment raw dialogs with retrieved relevant knowledge \cite{ghazvininejad2018knowledge,shuster-etal-2021-retrieval-augmentation,li2022exploring,DBLP:conf/coling/WangLWMZWL0022,zhang2022retgen}.
However, they all cover limited domains and knowledge types, questioning the transferability of pre-trained models to new domains and different types of knowledge.
Recently proposed GODEL \cite{peng2022godel} is the most related work to ours, which introduces an additional grounded pre-training phrase between general pre-training and fine-tuning, utilizing several grounded dialog corpora including question answering (grounded on passage), chit-chat (grounded on world knowledge), and task-oriented dialog (grounded on domain knowledge).
In contrast, our proposed pre-training method is more general without requiring such knowledge annotation and is demonstrated to be more effective in diverse grounded dialog generation tasks with different types of knowledge.

\subsection{Grounded Dialog Generation}
Grounding response generation on different types of knowledge is a vital ability for an informative, engaging, and reliable dialog agent.
Grounding knowledge could be common sense \cite{zhou2018commonsense}, persona \cite{zhang-etal-2018-personalizing}, world knowledge \cite{dinan2018wizard}, domain knowledge \cite{eric-etal-2017-key}, etc., in various forms ranging from structured knowledge such as dialog act \cite{wen-etal-2015-semantically}, knowledge graph \cite{zhou2018commonsense}, and table \cite{eric-etal-2017-key}, to unstructured knowledge such as keyword \cite{zhong2021keyword}, sentence \cite{Song2018AnEO}, and document \cite{dinan2018wizard}.
Since collecting natural knowledge-grounded conversation is often costly, learning a grounded dialog generation model with a few samples is a challenging but important topic.
However, the heterogeneity of knowledge leads to separated communities and specialized models \cite{zhou2018commonsense,wen-etal-2015-semantically,zhao2019low}, preventing the study of general grounded dialog generation models.
Recently, \citeauthor{UnifiedSKG} \shortcite{UnifiedSKG} propose to unify a series of structured knowledge grounded text generation tasks by serializing the input and output and applying the same text-to-text transformer.
Inspired by them, we unify the downstream tasks in a similar way.
But we focus on improving diverse grounded response generation tasks via pre-training and our proposed KPT does not need extra knowledge annotation.


\section{Method}
\subsection{Overview}
Given the dialog context $C_t=[U_1, U_2, ..., U_{t-1}]$, the goal of a knowledge-grounded dialog agent is to generate the next utterance $U_t$ grounded on knowledge $K_t$.
The knowledge $K_t$ could be persona descriptions \cite{zhang-etal-2018-personalizing}, dialog acts \cite{wen-etal-2015-semantically}, knowledge graph \cite{zhou2018commonsense}, etc.
Different grounding tasks may have different ways to retrieve relevant knowledge $K_t$ according to the context $C_t$, such as rules, domain APIs, and specialized knowledge retrievers.
Therefore, we separate the knowledge retrieval from knowledge-grounded response generation and focus on using pre-training to improve the latter (i.e. $P(U_t|C_t,K_t)$) where $K_t$ is already available.

To utilize large-scale dialog corpora without knowledge annotation for pre-training, we regard the keyword as the basic unit of knowledge and mine keywords from dialogs directly without any external knowledge sources or annotation efforts.
Inspired by \citeauthor{feng-etal-2021-language} \shortcite{feng-etal-2021-language}, we extract keywords in a dialog according to their LM loss judged by a pre-trained LM (\S \ref{sec:key}).
In \S \ref{sec:key_training}, we identify two kinds of knowledge usages on downstream tasks: 1) express all the knowledge in the response; 2) only select context-relevant knowledge, if any, to compose the response.
Thus, we devise two keyword-guided pre-training objectives that mimic the two kinds of knowledge usages respectively: generate the response according to the context and 1) keywords in the response or 2) randomly sampled keywords in the dialog, as illustrated in Table \ref{tab:example}.
Following \citeauthor{UnifiedSKG} \shortcite{UnifiedSKG} and \citeauthor{peng2022godel} \shortcite{peng2022godel},
we concatenate serialized knowledge with the context as the input and use T5 \cite{raffel2020t5} to generate the response for both pre-training and fine-tuning.

\subsection{Pre-training Datasets}

\begin{table}[t]
\small
\centering
\begin{tabular}{
    @{}
    >{\raggedright\arraybackslash}m{5.48em}
    >{\raggedleft\arraybackslash}m{3.2em}
    >{\raggedleft\arraybackslash}m{3.4em}
    >{\centering\arraybackslash}m{4.2em}
    >{\centering\arraybackslash}m{4.2em}
    @{}}
\toprule
Dataset      & Dialogs & Samples & Avg. Tok. & non-SW\% \\ \midrule
DailyDialog  & 11K     & 76K     & 11.2      & 46.4      \\
SGD          & 16K     & 309K    & 10.4      & 52.1      \\
Taskmaster-1 & 11K     & 211K    & 8.8       & 52.8      \\
Taskmaster-2 & 14K     & 218K    & 9.5       & 53.7      \\
Taskmaster-3 & 19K     & 360K    & 10.6      & 53.7      \\
MetaLWOZ     & 34K     & 323K    & 7.5       & 49.6      \\
Reddit       & 1.4M    & 5.2M    & 15.1      & 51.9      \\
WikiDialog   & 1.0M    & 4.7M    & 18.7      & 61.4      \\ \midrule
Total        & 2.5M    & 11.4M   &           &           \\ \bottomrule
\end{tabular}
\caption{Statistics of pre-training datasets. ``Avg. Tok." is the average tokens per turn. ``non-SW\%" is the average ratio of non-stop words in a turn. For WikiDialog, we only use system turns that come from wiki passages.}
\label{tab:pre-training-data}
\end{table}

As shown in Table \ref{tab:pre-training-data}, our pre-training datasets include DailyDialog \cite{li2017dailydialog}, Schema-Guided Dialog \cite{rastogi2020sgd}, Taskmaster-1/2/3 \cite{byrne-etal-2019-taskmaster,byrne-etal-2021-tickettalk}, MetaLWOZ \cite{li2020metalwoz}, DSTC8-Reddit \cite{lee2019dstc8}, and WikiDialog \cite{dai2022dialoginpainting}, covering chit-chats, goal-oriented dialogs, and information seeking dialogs.
We filter out the Reddit dialogs that contain URLs and dialogs that contain too long utterances ($>$256 tokens) in Reddit and WikiDialog.
WikiDialog is an artificial dataset containing 11M information-seeking dialogs where user utterances are generated by a model.
Therefore, we only use system turns for pre-training to ensure data quality.
Since WikiDialog is much larger than other pre-training datasets, we randomly choose 1M dialogs for pre-training to avoid an imbalanced training set.
Reddit and WikiDialog have comparable samples and constitute most of the pre-training data.
Nearly 50\% tokens are non-stop words across different datasets.

\begin{algorithm}[tb]
\caption{Prepare keyword-guided pre-training data}
\label{alg:algorithm}
\textbf{Input}: Dialog corpus $\mathcal{D}_{cov}$, Pre-trained LM $\phi$\\
\textbf{Parameter}: Keyword ratio $\alpha$\\
\textbf{Output}: Pre-training dataset $\mathcal{D}_{pre}$
\begin{algorithmic}[1] 
\STATE Initialize $\mathcal{D}_{pre}$ as an empty list\\
\FOR{dialog $D=U_1U_2...U_T$ in $\mathcal{D}_{cov}$}
\STATE Extract keywords as stated in \S \ref{sec:key} with $\phi$ and $\alpha$
\FOR{context-response pair $(C_t, U_t)$ in $D$}
\STATE $K_g\leftarrow KW_t$
\STATE Add $(C_t, K_g, U_t)$ to $\mathcal{D}_{pre}$
\STATE Initialize $K_n$ as an empty list
\STATE Sample random integer $j$ in range $[1,5]$
\STATE Sample $\min(j, T-1)$ turns from all $\{U_i\}_{i\neq t}$ and add their keywords $KW_i$ to $K_n$
\IF{$r \sim uniform(0, 1) < 0.8$}
\STATE Add $KW_t$ to $K_n$
\ENDIF
\STATE Add $(C_t, K_n, U_t)$ to $\mathcal{D}_{pre}$
\ENDFOR
\ENDFOR
\STATE \textbf{return} $\mathcal{D}_{pre}$
\end{algorithmic}
\end{algorithm}

\subsection{Keyword Extraction}
\label{sec:key}

\begin{table*}[t]
\centering
\small
\begin{tabular}{
    @{}
    >{\centering\arraybackslash}m{5em}|
    >{\raggedright\arraybackslash}m{13em}|
    >{\raggedright\arraybackslash}m{25em}|
    >{\raggedright\arraybackslash}m{9em}
    @{}}
\toprule
\textbf{Knowledge Source} & \multicolumn{1}{c|}{\textbf{Knowledge}} & \multicolumn{1}{c|}{\textbf{Input}} & \multicolumn{1}{c}{\textbf{Output}} \\ \midrule
Golden keywords (pre-train) &
  {[}{[}``CoD", ``hit"{]}, {[}``trying", ``become battlefield"{]}{]} &
  generate a response: \textbf{grounded knowledge: $|$ CoD : hit $|$ trying : become battlefield $|$} context: ... user: Yes, weapons now fire projectiles instead of hit scans. system: &
  But why, \textbf{CoD} has always been \textbf{hit} scan. Is CoD \textbf{trying} to \textbf{become battlefield}?\\ \midrule
Dialog acts (MultiWOZ) &
  \{``inform-hotel": \{``choice": ``9"\}, ``recommend-hotel": \{``name": ``Autumn House"\}\} &
  generate a response: \textbf{grounded knowledge: $|$ inform-hotel : choice : 9 $|$ recommend-hotel : name : Autumn House $|$} context: user: A guesthouse, please. Are there any in the cheap price range? system: &
  Oh, yes, there are \textbf{9}. I do recommend the \textbf{Autumn House}. \\ \midrule
Knowledge graph (OpenDialKG) &
  {[}{[}``Stranger in a Strange Land", ``has\_genre", ``Science Fiction"{]}{]} &
  generate a response: \textbf{grounded knowledge: $|$ Stranger in a Strange Land : has\_genre : Science Fiction $|$} context: user: I like Stranger in a Strange Land.  Can you recommend something for me? system: &
  Sure, do you like this book because it is the \textbf{genre} of \textbf{science fiction}?
   \\ \midrule \midrule
Noisy keywords (pre-train) &
  {[}{[}``Bullet travel"{]}, {[}``CoD", ``hit"{]}, {[}``trying", ``become battlefield"{]},...{]} &
  generate a response: \textbf{all knowledge: $|$ Bullet travel $|$ CoD : hit $|$ trying : become battlefield $|$ ... $|$} context: ... user: Yes, weapons now fire projectiles instead of hit scans. system: &
  But why, \textbf{CoD} has always been \textbf{hit} scan. Is CoD \textbf{trying} to \textbf{become battlefield}? \\ \midrule
Persona (PersonaChat) &
  {[}``i have one cat.", ``i am a kindergarten teacher.", ...{]} &
  generate a response: \textbf{all knowledge: $|$ i have one cat. $|$ i am a kindergarten teacher. $|$ ... $|$} context: ... user: I have a dog. his name is max. system: &
  That is cute I \textbf{have a cat}. \\ \midrule
Wikipedia (WoW) &
  {[}``Radiohead are an English rock band from Abingdon, Oxfordshire, formed in 1985.", ...{]} &
  generate a response: \textbf{all knowledge: $|$ Radiohead are an English rock band from Abingdon, Oxfordshire, formed in 1985. $|$ ... $|$} context: user: I just heard a song by Radiohead. It was great! system: &
  I love Radiohead! It's hard to believe that they've \textbf{been together since 1985}! \\ \bottomrule
\end{tabular}

\caption{
Examples for pre-training and downstream tasks, categorized into two groups according to how to use the knowledge.
In each row, we show the grounding knowledge, model input (serialized knowledge and context), and the golden response.
We use ``\textit{grounded knowledge}” to prompt the knowledge that should be faithfully grounded (the first three rows) and ``\textit{all knowledge}” to prompt the knowledge that should be selectively used (the last three rows).}
\label{tab:example}
\end{table*}

As there is no golden standard for keyword definition, we propose to extract words that convey information that could hardly be inferred from the context directly.
To this end, we select the most unpredictable words according to the context as the keywords.
Specifically, we concatenate the utterances $U_t=w^t_1 w^t_2...w^t_{|U_t|}$ in a dialog as one sequence $U_1 U_2...U_T$ and sort the words according to their conditional generation probability given by a pre-trained language model:
\begin{equation}
    P_{LM}(w^t_{i}|C_t,w^t_{<i})
\label{eq:lm}
\end{equation}
If a word is split into multiple tokens, its generation probability is the geometric average of these tokens' generation probabilities.
Then, we select the top $\alpha$ words ($\alpha$ is set to 30\% empirically) that have the lowest generation probability and are not stop words as keywords.
We ignore stop words since they often carry little information.
Finally, we merge adjacent keywords into ones and group the keywords by their belonging sentences to preserve the information of which keywords should be expressed together.
We denote keywords in a utterance $U$ that consists of $N$ sentences as $KW=\{\{k^1_1,k^1_2,...\},...,\{k^N_1,k^N_2,...\}\}$, where $k^i_{*}$ are keywords in the $i$-th sentence.


\subsection{Keyword-guided Pre-training}
\label{sec:key_training}
For each context-response pair $(C_t, U_t)$, we construct two kinds of knowledge inputs: golden keywords $K_g$ from the response $U_t$ and noisy keywords $K_n$ from the whole dialog $U_1U_2...U_T$, as shown in Algorithm \ref{alg:algorithm}.
The former simulates the knowledge that should be faithfully grounded on (e.g., dialog acts, relational triples) while the latter simulates the knowledge that should be selectively used (e.g., document, database, persona).
As illustrated in Table \ref{tab:example}, we use different prompts to distinguish these two kinds of knowledge.
Then we mix $(C_t, K_g, U_t)$ and $(C_t, K_n, U_t)$ in a ratio of 1:1 for keyword-guided response generation pre-training.

\paragraph{Response Generation with Golden Keywords}
This task is designed for the situation where the given knowledge must appear in the response.
We regard the keywords $KW_t$ extracted from the response $U_t$ as golden keywords $K_g$.
To serialize $K_g$, we use ``:" to join keywords in a sentence as a group and then use ``$|$" to join keyword groups, both in random order.
As shown in Table \ref{tab:example}, the input of the model is the concatenation of $K_g$ and $C_t$ while the output is $U_t$.
The loss function is:
\begin{equation}
    \mathcal{L}_{golden}(\theta)=-\sum^{|U_t|}_{i=1}\log p_\theta(w^t_i|C_t,w^t_{<i},K_g)
\end{equation}

\paragraph{Response Generation with Noisy Keywords}
In some cases \cite{zhang-etal-2018-personalizing,dinan2018wizard}, we only have a set of relevant knowledge instead of the exact grounding knowledge, where the model needs to decide whether to use knowledge and which knowledge to use.
To simulate these cases, we construct noisy keywords $K_n$ which consists of keywords from multiple turns in the same dialog to serve as ``negative" knowledge, as detailed in Algorithm \ref{alg:algorithm}.
To model the scenario where knowledge is not needed, $KW_t$ is only included in $K_n$ in 80\% of the time.
After mixing keyword groups from different turns, the serialization of $K_n$ is the same as $K_g$.
The loss function is:

\begin{equation}
    \mathcal{L}_{noisy}(\theta)=-\sum^{|U_t|}_{i=1}\log p_\theta(w^t_i|C_t,w^t_{<i},K_n)
\end{equation}


\section{Experimental Setup}
\subsection{Evaluation Datasets and Metrics}
We fine-tune models with KPT on four grounded dialog generation datasets covering various kinds of knowledge:

\begin{itemize}
    \item \textbf{MultiWOZ 2.1 (MWOZ)} \cite{budzianowski-etal-2018-multiwoz,eric-etal-2020-multiwoz} is a widely used multi-domain task-oriented dialog dataset.
    We conduct natural language generation (NLG) experiments on this dataset, where the response should strictly follow the given dialog acts.
    We treat each dialog act that consists of intent, slot, and value as a keyword group.
    Since the dialog acts already contain sufficient information, we use the latest three turns instead of the whole history as context to save computation.
    
    \item \textbf{OpenDialKG (ODKG)} \cite{moon-etal-2019-opendialkg} is an open-domain dialog dataset grounded on a knowledge graph (KG).
    Each turn is manually annotated with KG paths that link previously mentioned entities to the entities in this turn.
    For serialization, we treat each relational triple that consists of head entity, relation, and tail entity as a keyword group.
    Since the original data partition is not publicly available, we randomly split the data into training (70\%), validation (15\%), and test set (15\%).
    
    
    \item \textbf{PersonaChat (PC)} \cite{zhang-etal-2018-personalizing} is an open-domain dialog dataset where each speaker chats according to a pre-defined persona described by five sentences.
    We directly use ``$|$" to join all persona sentences as the knowledge input.
    
    \item \textbf{Wizard of Wikipedia (WoW)} \cite{dinan2018wizard} is an open-domain dialog dataset where system utterances are grounded on passages retrieved from Wikipedia.
    Each passage has a few sentences.
    Similar to PersonaChat, we use ``$|$" to join all sentences as the knowledge input.
    We merge the original seen and unseen test sets into one for evaluation.
\end{itemize}

Grounding knowledge of MultiWOZ and OpenDialKG is similar to the golden keywords in pre-training, while grounding knowledge of PersonaChat and WoW is similar to the noisy keywords.
Therefore, we use the corresponding prompt for each task and show the example in Table \ref{tab:example}.

We evaluate generated responses in two aspects: similarity to reference responses and knowledge utility.
For the former, we use perplexity (PPL), corpus-level BLEU-4 \cite{papineni-etal-2002-bleu,post-2018-call}, unigram F1 for non-stop words \cite{miller-etal-2017-parlai}, and Rouge-L \cite{lin-2004-rouge}.
When it comes to knowledge utility, it measures how much the model utilizes the provided grounding knowledge, and the metrics are dataset-specific.
For MultiWOZ, previous works \cite{wen-etal-2015-semantically} use Slot Error Rate (SER) to evaluate the ratio of missing and redundant slots in generated responses.
We use $1-$SER to be consistent with knowledge utility metrics of other datasets where bigger numbers are better.
For OpenDialKG, we evaluate micro entity F1 between generated responses and references, where a turn's candidate entities are the head and tail entities of grounding relational triples.
For PersonaChat and WoW, we calculate unigram F1 for non-stop words between generated responses and the concatenation of knowledge sentences.


\subsection{Baselines and Training Details}
Since our goal is to verify the effect of keyword-guided pre-training, we consider following baselines that are based on the same model architecture but pre-trained differently:
\begin{itemize}
    \item \textbf{T5} \cite{raffel2020t5}: Our backbone pre-trained model.
    \item \textbf{T5-RG}: We further pre-train T5 on our pre-training data to generate a response according to the context. It can be viewed as a T5 version of DialoGPT \cite{zhang-etal-2020-dialogpt}.
    \item \textbf{GODEL} \cite{peng2022godel}: Initialized by T5, GODEL is first trained with response generation objective on 147M Reddit dialogs extracted by \citet{zhang-etal-2020-dialogpt} and then trained with knowledge-grounded response generation objective on several datasets with knowledge annotation, including DSTC7 \cite{galley2019grounded}, MS MARCO \cite{nguyen2016ms}, UnifiedQA \cite{khashabi-etal-2020-unifiedqa}, and Schema-Guided Dialog \cite{rastogi2020sgd}.
\end{itemize}
In our main experiment, we use GPT-2 Large \cite{radford2019gpt2} for keyword extraction and perform keyword-guided pre-training (KPT) on T5 solely or after response generation pre-training (RG), denoted as \textbf{T5-KPT} and \textbf{T5-RG-KPT} respectively.
We use ConvLab-3 \cite{zhu2022convlab3} for dataset loading and model training.

We consider two sizes of model: 60M T5-small and 220M T5-base.
Since GODEL does not have a T5-small version, we only use the released T5-base version for comparison.
For both RG and KPT, we pre-train the models for 1 epoch.
The fine-tuning settings are the same for all models and downstream tasks.
We use the same knowledge serialization for all models.
During pre-training, we set the keyword ratio $\alpha$ to 0.3, and it is empirically chosen as illustrated in \S \ref{sec:5.3}.
To perform $K$-shot fine-tuning, we randomly sample $K$ dialogs from the original training and validation set respectively.
We alter $K$ in $\{50,100,200\}$ and sample the data 5 times for each $K$ in order to estimate the performance variance.
We fine-tune the models until the validation loss does not decrease for 5 consecutive epochs.
Models with the lowest validation losses during training are selected as the final models.
We use Adafactor optimizer with a constant learning rate 1e-3 for both pre-training and fine-tuning.
We set the batch size per GPU to 64 and use 8/2 Tesla V100 32G GPUs for pre-training/fine-tuning.
During inference, We adopt greedy decoding for a fair comparison of different models.

\section{Experiments and Analysis}
We organize our experiments as follows.
We compare keyword-guided pre-training with other pre-training methods in \S \ref{sec:5.1} and analyze the effect of each pre-training objective in Sec. \S \ref{sec:5.2}. 
We explore different keyword ratios and pre-trained LMs for keyword extraction in Sec. \S \ref{sec:5.3} and \S \ref{sec:5.4}.
Finally, we conduct human evaluation in \S \ref{sec:5.5}.
Due to the space limit, we place completed experimental results in the appendix for reference.

\subsection{Main Experiment}
\label{sec:5.1}
\begin{table*}[t]
\small
\centering
\begin{tabular}{llcccccccccccc}
\toprule
\multirow{2}{*}{Dataset} &
  \multirow{2}{*}{Model} &
  \multicolumn{3}{c}{BLEU-4} &
  \multicolumn{3}{c}{Unigram F1} &
  \multicolumn{3}{c}{Rouge-L} &
  \multicolumn{3}{c}{Knowledge Utility} \\ \cmidrule(lr){3-5} \cmidrule(lr){6-8} \cmidrule(lr){9-11} \cmidrule(lr){12-14}  
                             &           & 50   & 100  & 200          & 50   & 100  & 200           & 50   & 100  & 200           & 50            & 100  & 200           \\ \midrule
\multirow{5}{*}{MultiWOZ}    & T5        & 16.7 & 20.8 & 26.7         & 35.0 & 45.1 & 50.3          & 33.2 & 40.8 & 45.7          & 65.9          & 82.0 & 88.2          \\
                             & GODEL     & 17.8 & 21.5 & 26.1         & 38.3 & 44.1 & 48.9          & 36.0 & 40.3 & 44.7          & 69.5          & 82.4 & 88.3          \\
                             & T5-RG     & 17.4 & 20.9 & 25.5         & 36.2 & 41.7 & 47.8          & 35.1 & 39.2 & 44.3          & 64.2          & 75.0 & 81.9          \\ \cmidrule{2-14}
 &
  T5-KPT &
  23.1 &
  \textbf{27.1} &
  28.7 &
  \textbf{46.2} &
  \textbf{51.0} &
  53.4 &
  41.8 &
  \textbf{45.6} &
  47.9 &
  84.0 &
  \textbf{91.8} &
  93.7 \\
 &
  T5-RG-KPT &
  \textbf{23.4} &
  26.8 &
  \textbf{29.7} &
  \textbf{46.2} &
  50.9 &
  \textbf{54.3} &
  \textbf{42.0} &
  45.4 &
  \textbf{48.5} &
  \textbf{84.2} &
  91.7 &
  \textbf{95.0} \\ \midrule
\multirow{5}{*}{OpenDialKG}  & T5        & 7.9  & 7.3  & 7.5          & 25.0 & 23.1 & 23.1          & 23.0 & 23.1 & 23.8          & 54.6          & 57.2 & 52.9          \\
                             & GODEL     & 9.4  & 9.9  & 11.0         & 27.8 & 33.1 & 31.8          & 27.4 & 30.7 & 29.7          & 59.9          & 68.4 & 68.4          \\
                             & T5-RG     & 7.8  & 10.1 & 11.0         & 23.3 & 28.0 & 29.9          & 24.5 & 28.2 & 29.6          & 52.6          & 60.8 & 65.1          \\ \cmidrule{2-14}
 &
  T5-KPT &
  \textbf{9.7} &
  \textbf{12.5} &
  \textbf{13.2} &
  \textbf{30.4} &
  \textbf{34.8} &
  \textbf{36.5} &
  \textbf{28.9} &
  \textbf{31.9} &
  \textbf{33.8} &
  \textbf{63.6} &
  \textbf{72.9} &
  \textbf{74.0} \\
                             & T5-RG-KPT & 8.3  & 12.1 & 12.4         & 27.5 & 34.7 & 35.2          & 27.2 & 31.8 & 32.9          & 57.7          & 71.3 & 72.0          \\ \midrule
\multirow{5}{*}{PersonaChat} & T5        & 1.5  & 2.2  & 2.5          & 5.9  & 5.2  & 7.4           & 15.2 & 16.4 & 17.3          & 11.9          & 11.8 & 17.5          \\
                             & GODEL     & 2.5  & 2.8  & 3.6          & 7.9  & 7.9  & 8.9           & 17.2 & 17.6 & 18.6          & 16.0          & 16.3 & 18.7          \\
                             & T5-RG     & 2.8  & 3.1  & 3.7          & 7.3  & 8.3  & 9.0           & 17.7 & 18.9 & 19.1          & 14.2          & 13.9 & 18.1          \\ \cmidrule{2-14}
 &
  T5-KPT &
  \textbf{2.9} &
  3.3 &
  3.5 &
  \textbf{8.8} &
  8.9 &
  9.3 &
  \textbf{18.4} &
  18.7 &
  19.0 &
  \textbf{20.1} &
  \textbf{18.4} &
  19.5 \\
 &
  T5-RG-KPT &
  \textbf{2.9} &
  \textbf{3.7} &
  \textbf{3.8} &
  8.6 &
  \textbf{9.1} &
  \textbf{9.6} &
  \textbf{18.4} &
  \textbf{19.3} &
  \textbf{19.3} &
  19.1 &
  17.1 &
  \textbf{20.8} \\ \midrule
\multirow{5}{*}{WoW}         & T5        & 4.0  & 2.3  & 3.1          & 12.0 & 8.9  & 9.4           & 16.1 & 14.0 & 14.9          & \textbf{17.1} & 7.7  & 9.7           \\
                             & GODEL     & 2.8  & 4.1  & 4.5          & 9.4  & 11.6 & 12.7          & 15.3 & 17.1 & 17.5          & 6.8           & 10.6 & \textbf{13.6} \\
                             & T5-RG     & 2.5  & 3.8  & 4.5          & 9.4  & 11.3 & 12.3          & 15.5 & 17.0 & 17.8          & 5.1           & 8.1  & 10.0          \\ \cmidrule{2-14}
 &
  T5-KPT &
  \textbf{5.3} &
  \textbf{4.8} &
  5.1 &
  \textbf{13.7} &
  \textbf{13.1} &
  \textbf{13.5} &
  \textbf{18.2} &
  \textbf{17.9} &
  18.4 &
  14.0 &
  \textbf{12.4} &
  12.6 \\
                             & T5-RG-KPT & 5.2  & 4.7  & \textbf{5.2} & 13.1 & 12.8 & \textbf{13.5} & 17.6 & 17.6 & \textbf{18.5} & 13.2          & 11.7 & 12.5          \\ \bottomrule
\end{tabular}
\caption{Performance of models based on T5-base. 
Results of 50/100/200-shot are shown in different columns.
We report the means of 5 runs and move the standard deviations to the appendix to save space. The best results are in bold.}
\label{tab:t5base}
\end{table*}

Table \ref{tab:t5base} and \ref{tab:t5small} show the few-shot fine-tuning results of models based on T5-base and T5-small respectively.
Comparing T5 and T5-RG, We can see that the response generation pre-training does not always bring benefits for grounded dialog generation.
T5-RG is worse than T5 on MultiWOZ and OpenDialKG, where the response should be faithfully grounded on the given knowledge.
We contend that RG pre-training teaches the model to generate the response based on the context and therefore tends to ignore the (heterogeneous) grounding knowledge.
On the contrary, T5-KPT and T5-RG-KPT outperform T5 and T5-RG in most cases, especially w.r.t. the knowledge utility score, demonstrating the effectiveness of our proposed KPT on grounded dialog generation tasks.
T5-KPT is comparable to T5-RG-KPT, indicating that RG pre-training before KPT is not necessary for grounded dialog generation.
Nevertheless, we use T5-RG-KPT for subsequent analysis since RG pre-training may handle some scenarios where knowledge is not needed.
GODEL outperforms T5 in most settings but it is still inferior to KPT.
Although GODEL has been trained on several knowledge-grounded dialog datasets, the specific types of knowledge of these datasets may limit the transfer effect to different downstream knowledge grounding tasks.

\subsection{Pre-training Objectives Ablation}
\label{sec:5.2}
\begin{table}[t]
\small
\centering
\begin{tabular}{@{}llcc@{}}
\toprule
Dataset               & Model       & Unigram F1              & Knowledge Utility       \\ \midrule
\multirow{6}{*}{MWOZ} & T5          & 38.0/44.9/49.7          & 77.1/83.3/88.9          \\
                      & T5-RG       & 36.5/41.7/47.4          & 67.2/76.6/82.5          \\ \cmidrule(l){2-4} 
                      & T5-KPT      & 47.1/50.7/53.6          & 89.5/91.8/94.6          \\
                      & T5-RG-KPT   & 46.8/50.6/53.0          & 88.0/92.0/94.0          \\
                      & T5-RG-$K_g$ & \textbf{48.0/50.7/52.5} & \textbf{91.7/92.8/93.8} \\
                      & T5-RG-$K_n$ & 41.7/46.7/51.0          & 78.8/83.9/88.5          \\ \midrule
\multirow{6}{*}{ODKG} & T5          & 20.4/25.6/27.9          & 46.7/58.0/61.5          \\
                      & T5-RG       & 21.3/26.1/29.6          & 47.1/57.8/63.5          \\ \cmidrule(l){2-4} 
                      & T5-KPT      & 32.4/35.1/34.9          & 67.6/72.2/71.2          \\
                      & T5-RG-KPT   & 32.7/35.2/34.6          & 69.4/72.9/70.9          \\
                      & T5-RG-$K_g$ & 32.7/\textbf{33.9/35.0} & 68.4/\textbf{70.5/71.1} \\
                      & T5-RG-$K_n$ & \textbf{33.8}/33.7/33.8          & \textbf{70.1}/70.0/69.1          \\ \midrule
\multirow{6}{*}{PC}   & T5          & 4.0/4.5/5.6             & 15.1/17.1/17.4          \\
                      & T5-RG       & 7.2/7.3/7.5             & 13.9/14.6/16.5          \\ \cmidrule(l){2-4} 
                      & T5-KPT      & 7.9/7.8/8.0             & 22.2/19.8/18.2          \\
                      & T5-RG-KPT   & 8.1/7.9/8.2             & 18.1/20.0/19.1          \\
                      & T5-RG-$K_g$ & 7.7/7.4/7.5             & \textbf{21.5}/14.4/17.9          \\
                      & T5-RG-$K_n$ & \textbf{7.9/7.9/8.2}    & 15.7/\textbf{18.1/19.0} \\ \midrule
\multirow{6}{*}{WoW}  & T5          & 10.1/9.0/8.6            & 13.7/11.0/10.5          \\ 
                      & T5-RG       & 9.3/10.0/10.6           & 7.1/8.4/9.6             \\ \cmidrule(l){2-4}
                      & T5-KPT      & 13.2/11.7/12.2          & 15.1/10.0/12.1          \\
                      & T5-RG-KPT   & 13.3/12.2/13.1          & 15.2/11.8/13.1          \\
                      & T5-RG-$K_g$ & 12.3/11.0/12.1          & \textbf{14.7}/10.1/\textbf{11.4}         \\
                      & T5-RG-$K_n$ & \textbf{12.5/11.6/12.3} & 12.9/\textbf{10.3/11.4} \\ \bottomrule
\end{tabular}
\caption{Performance of models based on T5-small. 
Results of 50/100/200-shot are separated by a slash.
The best results among T5-RG-$K_g$ and T5-RG-$K_n$ are in bold.}
\label{tab:t5small}
\end{table}

To investigate the effect of each pre-training objective, we pre-train the grounded response generation model using either golden keywords $K_g$ or noisy keywords $K_n$ after RG pre-training, and they are referred to as T5-RG-$K_g$ and T5-RG-$K_n$ respectively.
Results of different model ablations of T5-small model size are shown in Table \ref{tab:t5small}.
We could see that these two kinds of keywords contribute differently.
On MultiWOZ, the superior performance of T5-RG-KPT mainly comes from $K_g$, while $K_n$ contributes more on PersonaChat.
This observation is also intuitive as MultiWOZ needs faithfully grounding on the knowledge the most, while PersonaChat needs selectively grounding the most.
On WoW, these two types of keywords have similar contributions to T5-RG-KPT on top of T5-RG.
On OpenDialKG, their influences are not clear, since the relative performances differ across shots.

\subsection{Keyword Ratio}
\label{sec:5.3}
\begin{figure}[t]
\centering
\includegraphics[width=\columnwidth]{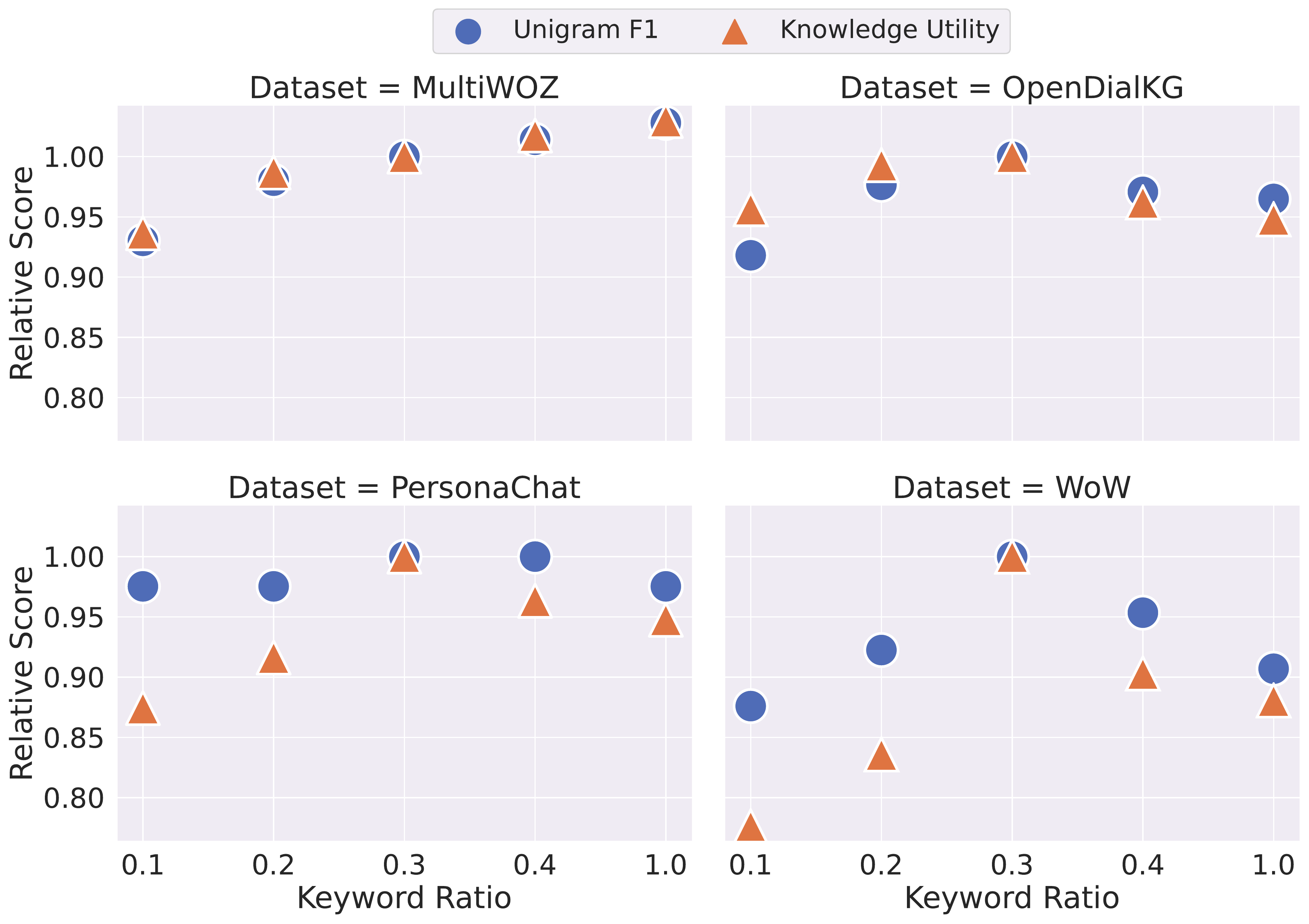} 
\caption{Performance of models based on T5-small with different keyword ratios ($\alpha$). 
We average 50, 100, and 200 shots' performance and report the relative scores compared to the setting of $\alpha$=0.3.
}
\label{fig:keyratio}
\end{figure}

In this section, we investigate the influence of keyword ratio in KPT.
As we mentioned before, keywords are chosen from non-stop words (roughly 50\% of all words) to serve as pseudo grounding knowledge.
Therefore, we vary the keyword ratio $\alpha$ in $\{0.1, 0.2, 0.3, 0.4, 1.0\}$, where 1.0 means we use all non-stop words.
When $\alpha$ is 0, KPT degenerates to RG.
Larger $\alpha$ leaks more information, which encourages the model to ground on the knowledge more.
However, too large $\alpha$ will make the model ignore the context.
As shown in Figure \ref{fig:keyratio}, as $\alpha$ increases, the performance on all tasks except MultiWOZ first increases until $\alpha$=0.3 then decreases, demonstrating the necessity of choosing appropriate $\alpha$.
Performance on MultiWOZ continually increases with larger $\alpha$, which can be attributed to the fact that this task is heavily grounded on knowledge and rarely rely on context.

\subsection{Keyword Extraction Model}
\label{sec:5.4}
We also explore different keyword extraction methods in KPT.
Besides GPT-2 Large (774M), we also try DialoGPT Large (762M) as the keyword extraction model.
We compare the results of these two keyword extraction models and a vanilla random extraction approach in Table \ref{tab:ke} for T5-RG-KPT of T5-small model size.
Compared with using GPT-2, random extraction leads to worse performance on PersonaChat and WoW, while using DialoGPT performs poorly on OpenDialKG and PersonaChat.
This result shows the influence of bias in keyword extraction on the final performance of the grounded response generation task.
We assume that GPT-2 may create a more accurate estimation of Eq. (\ref{eq:lm}) on diverse keyword-guided pre-training corpora than DialoGPT since it is pre-trained on more diverse text corpora.
Nevertheless, we also found that the performance gaps between different keyword extraction methods are not significant, and using random extraction can already achieve satisfying performance.

\begin{table}[t]
\small
\centering
\begin{tabular}{@{}llcc@{}}
\toprule
Dataset               & KE Model    & Unigram F1     & Knowledge Utility \\ \midrule
\multirow{3}{*}{MWOZ} & GPT-2    & 46.8/50.6/53.0 & 88.0/\textbf{92.0}/94.0    \\
                      & DialoGPT & \textbf{48.1}/51.0/53.0 & \textbf{89.8}/91.7/94.3    \\
                      & Random   & 47.1/\textbf{51.1}/\textbf{54.0} & 87.3/91.8/\textbf{94.7}    \\ \midrule
\multirow{3}{*}{ODKG} & GPT-2    & \textbf{32.7/35.2/34.6} & \textbf{69.4/72.9/70.9}    \\
                      & DialoGPT & 29.6/32.8/33.4 & 64.6/69.4/69.6    \\
                      & Random   & 32.6/34.6/33.8 & 68.4/72.2/69.4    \\ \midrule
\multirow{3}{*}{PC}   & GPT-2    & \textbf{8.1/7.9/8.2}    & \textbf{18.1/20.0/19.1}    \\
                      & DialoGPT & 7.8/7.6/8.0    & 14.1/18.3/17.0    \\
                      & Random   & 7.7/7.7/8.0    & 13.5/19.4/18.0    \\ \midrule
\multirow{3}{*}{WoW}  & GPT-2    & \textbf{13.3/12.2/13.1} & \textbf{15.2}/11.8/\textbf{13.1}    \\
                      & DialoGPT & 11.8/12.1/12.7 & 12.2/\textbf{11.9}/13.0    \\
                      & Random   & 12.1/11.6/11.9 & 12.2/10.6/11.0    \\ \bottomrule
\end{tabular}
\caption{
Performance of T5-RG-KPT models based on T5-small but with different keyword extraction models.
Results of 50/100/200-shot are separated by a slash, and best results are in bold.}
\label{tab:ke}
\end{table}

\subsection{Human Evaluation}
\label{sec:5.5}
We further employed workers on Amazon Mechanical Turk to evaluate the responses generated by T5-RG-KPT and two baselines T5-RG and GODEL.
All models are of T5-base model size and fine-tuned on 50 dialogs.
For a sample, we asked 3 workers to compare 2 responses (ours and the baseline model's) and select the better one w.r.t. context coherence and knowledge utility, resulting in 4 (tasks) * 2 (baselines) * 2 (aspects) * 100 (samples) * 3 (workers) = 4800 scores. 141 workers attended the human evaluation in total.

\begin{table}[]
\centering
\begin{tabular}{@{}lll@{}}
\toprule
Win Rate       & Our vs T5-RG   & Our vs GODEL   \\ \midrule
MWOZ-Coherence & 35.3 vs 34.7   & 42.7 vs 31.7   \\
MWOZ-Knowledge & 50.3 vs 22.7** & 46.0 vs 24.0** \\ \midrule
ODKG-Coherence & 49.3 vs 40.3*  & 48.7 vs 41.3*  \\
ODKG-Knowledge & 52.3 vs 39.0   & 44.7 vs 40.7   \\ \midrule
PC-Coherence   & 43.7 vs 40.0   & 49.0 vs 37.0   \\
PC-Knowledge   & 47.3 vs 40.3   & 41.3 vs 39.7** \\ \midrule
WoW-Coherence  & 48.3 vs 44.0   & 47.3 vs 44.7   \\
WoW-Knowledge  & 54.0 vs 39.3   & 48.0 vs 44.7   \\ \bottomrule
\end{tabular}
\caption{
Pairwise human evaluation. * means Fleiss' kappa $\kappa$\textless 0 (poor agreement) and ** means $\kappa$\textgreater 0.2 (fair agreement), while others are within [0, 0.2] (slight agreement).}
\label{tab:human_eval}
\end{table}

The result is shown in Table \ref{tab:human_eval}.
Since dialog evaluation is rather subjective and subtle, the low agreement is acceptable.
We can see that T5-RG-KPT is better than T5-RG and GODEL, consistent with the automatic results. Compared with T5-RG, our model mostly improve knowledge utility, while compared with GODEL, the improvement in context coherence is more obvious.

\section{Conclusion}
In this paper, we propose \textit{keyword-guided pre-training} (KPT), a novel dialog pre-training method for grounded dialog generation.
KPT does not need the dialogs to be annotated with grounding knowledge and outperforms standard response generation pre-training and supervised knowledge-grounded pre-training on several downstream tasks with diverse knowledge sources.
Further analysis shows that our two pre-training objectives complement each other, enabling KPT to handle scenarios that make use of knowledge at different levels.
Altogether, we contend that KPT is an effective, scalable, and versatile pre-training technique towards build better grounded dialog systems.

\section*{Acknowledgement}

This work was supported by the National Key Research and Development Program of China (No. 2021ZD0113304), the National Science Foundation for Distinguished Young Scholars (with No. 62125604), the NSFC projects (Key project with No. 61936010 and regular project with No. 61876096), and sponsored by Tsinghua-Toyota Joint Research Fund.

\bibliography{custom,anthology}

\appendix

\begin{table*}[t]
\small
\centering
\begin{tabular}{@{}llcccc@{}}
\toprule
Dataset               & Model       &  BLEU-4 & Unigram F1 & Rouge-L      & Knowledge Utility       \\ \midrule
\multirow{5}{*}{MWOZ} &
T5 &
  $16.7_{2.1}$/$20.8_{3.0}$/$26.7_{1.7}$ &
  $35.0_{3.9}$/$45.1_{3.3}$/$50.3_{1.0}$ &
  $33.2_{2.1}$/$40.8_{2.8}$/$45.7_{0.5}$ &
  $65.9_{8.2}$/$82.0_{3.8}$/$88.2_{5.2}$ \\
 &
  GODEL &
  $17.8_{2.9}$/$21.5_{1.5}$/$26.1_{2.5}$ &
  $38.3_{3.6}$/$44.1_{1.6}$/$48.9_{2.4}$ &
  $36.0_{2.9}$/$40.3_{1.3}$/$44.7_{1.4}$ &
  $69.5_{9.8}$/$82.4_{2.7}$/$88.3_{4.8}$ \\
 &
  T5-RG &
  $17.4_{1.0}$/$20.9_{1.6}$/$25.5_{2.6}$ &
  $36.2_{2.5}$/$41.7_{1.9}$/$47.8_{3.5}$ &
  $35.1_{1.5}$/$39.2_{1.6}$/$44.3_{2.8}$ &
  $64.2_{8.3}$/$75.0_{3.5}$/$81.9_{8.2}$ \\ \cmidrule(l){2-6} 
 &
  T5-KPT &
  $23.1_{1.5}$/$27.1_{1.6}$/$28.7_{1.5}$ &
  $46.2_{2.6}$/$51.0_{1.5}$/$53.4_{1.2}$ &
  $41.8_{1.8}$/$45.6_{0.8}$/$47.9_{1.3}$ &
  $84.0_{4.8}$/$91.8_{2.7}$/$93.7_{2.3}$ \\
 &
  T5-RG-KPT &
  $23.4_{0.7}$/$26.8_{1.2}$/$29.7_{0.7}$ &
  $46.2_{0.5}$/$50.9_{1.2}$/$54.3_{1.6}$ &
  $42.0_{0.8}$/$45.4_{1.0}$/$48.5_{1.4}$ &
  $84.2_{3.2}$/$91.7_{1.8}$/$95.0_{1.2}$ \\ \midrule
\multirow{5}{*}{ODKG} &
  T5 &
  $7.9_{3.6}$/$7.3_{2.0}$/$7.5_{0.9}$ &
  $25.0_{9.4}$/$23.1_{3.8}$/$23.1_{3.3}$ &
  $23.0_{6.3}$/$23.1_{3.1}$/$23.8_{2.6}$ &
  $54.6_{23.9}$/$57.2_{7.8}$/$52.9_{4.4}$ \\
 &
  GODEL &
  $9.4_{1.9}$/$9.9_{1.4}$/$11.0_{1.4}$ &
  $27.8_{4.2}$/$33.1_{2.3}$/$31.8_{3.3}$ &
  $27.4_{2.5}$/$30.7_{1.8}$/$29.7_{2.3}$ &
  $59.9_{9.3}$/$68.4_{2.8}$/$68.4_{4.1}$ \\
 &
  T5-RG &
  $7.8_{0.5}$/$10.1_{0.8}$/$11.0_{1.9}$ &
  $23.3_{1.7}$/$28.0_{1.1}$/$29.9_{2.7}$ &
  $24.5_{1.2}$/$28.2_{0.5}$/$29.6_{1.9}$ &
  $52.6_{2.8}$/$60.8_{3.7}$/$65.1_{6.2}$ \\ \cmidrule(l){2-6} 
 &
  T5-KPT &
  $9.7_{2.1}$/$12.5_{0.9}$/$13.2_{0.6}$ &
  $30.4_{3.8}$/$34.8_{1.9}$/$36.5_{1.5}$ &
  $28.9_{2.2}$/$31.9_{0.8}$/$33.8_{0.6}$ &
  $63.6_{7.5}$/$72.9_{2.8}$/$74.0_{1.8}$ \\
 &
  T5-RG-KPT &
  $8.3_{3.4}$/$12.1_{1.3}$/$12.4_{0.9}$ &
  $27.5_{5.8}$/$34.7_{2.0}$/$35.2_{2.3}$ &
  $27.2_{3.3}$/$31.8_{1.0}$/$32.9_{1.2}$ &
  $57.7_{10.8}$/$71.3_{3.3}$/$72.0_{2.8}$ \\ \midrule
\multirow{5}{*}{PC} &
  T5 &
  $1.5_{0.7}$/$2.2_{0.6}$/$2.5_{0.5}$ &
  $5.9_{1.1}$/$5.2_{1.6}$/$7.4_{0.6}$ &
  $15.2_{2.2}$/$16.4_{0.3}$/$17.3_{0.2}$ &
  $11.9_{1.3}$/$11.8_{6.8}$/$17.5_{3.5}$ \\
 &
  GODEL &
  $2.5_{0.5}$/$2.8_{0.5}$/$3.6_{0.1}$ &
  $7.9_{1.6}$/$7.9_{1.6}$/$8.9_{0.7}$ &
  $17.2_{0.4}$/$17.6_{1.1}$/$18.6_{0.2}$ &
  $16.0_{3.9}$/$16.3_{7.2}$/$18.7_{2.9}$ \\
 &
  T5-RG &
  $2.8_{0.2}$/$3.1_{0.4}$/$3.7_{0.3}$ &
  $7.3_{0.9}$/$8.3_{0.6}$/$9.0_{0.5}$ &
  $17.7_{0.7}$/$18.9_{0.5}$/$19.1_{0.4}$ &
  $14.2_{1.3}$/$13.9_{3.5}$/$18.1_{2.4}$ \\ \cmidrule(l){2-6} 
 &
  T5-KPT &
  $2.9_{0.4}$/$3.3_{0.3}$/$3.5_{0.3}$ &
  $8.8_{0.5}$/$8.9_{0.6}$/$9.3_{0.6}$ &
  $18.4_{0.7}$/$18.7_{0.4}$/$19.0_{0.7}$ &
  $20.1_{6.3}$/$18.4_{2.9}$/$19.5_{4.9}$ \\
 &
  T5-RG-KPT &
  $2.9_{0.2}$/$3.7_{0.3}$/$3.8_{0.1}$ &
  $8.6_{0.7}$/$9.1_{0.7}$/$9.6_{0.6}$ &
  $18.4_{0.5}$/$19.3_{0.4}$/$19.3_{0.4}$ &
  $19.1_{9.9}$/$17.1_{2.8}$/$20.8_{5.3}$ \\ \midrule
\multirow{5}{*}{WoW} &
  T5 &
  $4.0_{0.8}$/$2.3_{0.6}$/$3.1_{1.1}$ &
  $12.0_{0.2}$/$8.9_{2.0}$/$9.4_{1.7}$ &
  $16.1_{0.3}$/$14.0_{1.2}$/$14.9_{1.5}$ &
  $17.1_{3.3}$/$7.7_{2.7}$/$9.7_{3.9}$ \\
 &
  GODEL &
  $2.8_{0.5}$/$4.1_{0.6}$/$4.5_{0.8}$ &
  $9.4_{1.8}$/$11.6_{0.7}$/$12.7_{2.0}$ &
  $15.3_{1.5}$/$17.1_{0.6}$/$17.5_{1.3}$ &
  $6.8_{3.6}$/$10.6_{2.7}$/$13.6_{2.5}$ \\
 &
  T5-RG &
  $2.5_{0.1}$/$3.8_{0.3}$/$4.5_{0.6}$ &
  $9.4_{0.3}$/$11.3_{0.7}$/$12.3_{0.8}$ &
  $15.5_{0.3}$/$17.0_{0.3}$/$17.8_{0.5}$ &
  $5.1_{0.6}$/$8.1_{0.9}$/$10.0_{1.7}$ \\ \cmidrule(l){2-6} 
 &
  T5-KPT &
  $5.3_{0.5}$/$4.8_{0.9}$/$5.1_{0.6}$ &
  $13.7_{0.4}$/$13.1_{1.2}$/$13.5_{0.9}$ &
  $18.2_{0.3}$/$17.9_{0.9}$/$18.4_{0.7}$ &
  $14.0_{1.4}$/$12.4_{2.9}$/$12.6_{2.3}$ \\
 &
  T5-RG-KPT &
  $5.2_{0.5}$/$4.7_{1.1}$/$5.2_{0.5}$ &
  $13.1_{1.1}$/$12.8_{1.5}$/$13.5_{0.2}$ &
  $17.6_{0.9}$/$17.6_{0.9}$/$18.5_{0.4}$ &
  $13.2_{1.7}$/$11.7_{3.8}$/$12.5_{2.1}$ \\ \bottomrule
\end{tabular}
\caption{Performance of models based on T5-base. 
Results of 50/100/200-shot are separated by a slash. 
We report the means and standard deviations of 5 runs.}
\label{tab:t5base_all}
\end{table*}

\begin{table*}[t]
\small
\centering
\begin{tabular}{@{}llcccc@{}}
\toprule
Dataset               & Model       &  BLEU-4 & Unigram F1 & Rouge-L      & Knowledge Utility       \\ \midrule
\multirow{12}{*}{MWOZ} &
  T5 &
  $17.4_{1.8}$/$21.9_{2.7}$/$26.5_{1.8}$ &
  $38.0_{4.9}$/$44.9_{2.7}$/$49.7_{1.4}$ &
  $34.2_{3.4}$/$40.2_{2.6}$/$44.7_{1.2}$ &
  $77.1_{9.6}$/$83.3_{5.1}$/$88.9_{2.8}$ \\
 &
  T5-RG &
  $17.4_{0.5}$/$21.2_{1.2}$/$25.6_{0.7}$ &
  $36.5_{0.8}$/$41.7_{1.2}$/$47.4_{1.6}$ &
  $35.1_{0.7}$/$39.3_{1.1}$/$44.0_{0.9}$ &
  $67.2_{4.2}$/$76.6_{5.0}$/$82.5_{3.1}$ \\ \cmidrule(l){2-6} 
 &
  T5-KPT &
  $23.8_{0.8}$/$26.0_{1.2}$/$29.1_{0.6}$ &
  $47.1_{1.0}$/$50.7_{1.5}$/$53.6_{0.9}$ &
  $41.8_{0.7}$/$44.8_{1.2}$/$47.6_{0.7}$ &
  $89.5_{2.2}$/$91.8_{1.5}$/$94.6_{0.7}$ \\
 &
  T5-RG-KPT &
  $23.3_{0.6}$/$26.3_{1.1}$/$29.0_{0.4}$ &
  $46.8_{1.1}$/$50.6_{1.2}$/$53.0_{0.9}$ &
  $41.5_{0.6}$/$44.7_{0.8}$/$47.1_{0.6}$ &
  $88.0_{2.9}$/$92.0_{1.6}$/$94.0_{1.7}$ \\
 &
  T5-RG-$K_g$ &
  $24.1_{0.5}$/$26.8_{0.7}$/$29.0_{0.6}$ &
  $48.0_{0.9}$/$50.7_{0.9}$/$52.5_{1.0}$ &
  $42.1_{0.4}$/$44.9_{0.8}$/$46.8_{1.2}$ &
  $91.7_{1.9}$/$92.8_{1.4}$/$93.8_{1.1}$ \\
 &
  T5-RG-$K_n$ &
  $19.8_{1.9}$/$24.4_{1.1}$/$28.0_{0.4}$ &
  $41.7_{1.7}$/$46.7_{1.4}$/$51.0_{0.7}$ &
  $38.5_{1.4}$/$42.7_{1.3}$/$46.5_{0.7}$ &
  $78.8_{3.1}$/$83.9_{1.5}$/$88.5_{1.9}$ \\ \cmidrule(l){2-6} 
 &
  T5-RG-KPT(DGPT) &
  $23.5_{1.0}$/$26.7_{0.7}$/$29.3_{0.6}$ &
  $48.1_{0.6}$/$51.0_{0.8}$/$53.0_{0.8}$ &
  $42.2_{0.6}$/$45.2_{0.5}$/$47.1_{0.6}$ &
  $89.8_{1.6}$/$91.7_{0.9}$/$94.3_{1.2}$ \\
 &
  T5-RG-KPT(Rand) &
  $23.3_{0.8}$/$26.6_{0.5}$/$29.6_{0.4}$ &
  $47.1_{1.7}$/$51.1_{0.6}$/$54.0_{0.6}$ &
  $42.0_{1.3}$/$45.3_{0.4}$/$48.0_{0.8}$ &
  $87.3_{2.2}$/$91.8_{0.7}$/$94.7_{1.0}$ \\ \cmidrule(l){2-6} 
 &
  T5-RG-KPT($\alpha$=0.1) &
  $20.5_{0.7}$/$23.8_{1.0}$/$27.4_{0.4}$ &
  $42.5_{0.7}$/$46.5_{1.4}$/$50.7_{0.8}$ &
  $38.9_{0.6}$/$42.4_{1.3}$/$45.8_{0.7}$ &
  $81.6_{4.8}$/$85.5_{1.7}$/$89.4_{1.3}$ \\
 &
  T5-RG-KPT($\alpha$=0.2) &
  $21.9_{1.0}$/$25.5_{1.5}$/$28.1_{0.2}$ &
  $45.4_{0.5}$/$49.4_{1.0}$/$52.5_{0.7}$ &
  $40.3_{0.6}$/$44.0_{0.7}$/$46.7_{0.5}$ &
  $87.2_{3.6}$/$90.0_{2.3}$/$93.1_{1.2}$ \\
 &
  T5-RG-KPT($\alpha$=0.4) &
  $23.7_{0.9}$/$27.0_{0.7}$/$29.0_{0.6}$ &
  $47.9_{0.7}$/$51.3_{1.1}$/$53.2_{0.7}$ &
  $42.2_{0.5}$/$45.2_{0.7}$/$47.1_{0.6}$ &
  $90.4_{1.1}$/$93.3_{1.5}$/$94.9_{0.7}$ \\
 &
  T5-RG-KPT($\alpha$=1.0) &
  $24.2_{0.6}$/$26.9_{0.3}$/$29.4_{0.4}$ &
  $48.7_{0.9}$/$51.5_{0.7}$/$54.2_{0.6}$ &
  $42.7_{0.6}$/$45.4_{0.6}$/$47.8_{0.4}$ &
  $92.9_{1.1}$/$93.5_{0.9}$/$95.7_{0.5}$ \\ \midrule
\multirow{12}{*}{ODKG} &
  T5 &
  $6.8_{2.8}$/$8.9_{2.2}$/$10.0_{1.7}$ &
  $20.4_{5.6}$/$25.6_{4.3}$/$27.9_{2.3}$ &
  $21.0_{3.9}$/$24.4_{3.8}$/$27.3_{1.6}$ &
  $46.7_{14.3}$/$58.0_{9.4}$/$61.5_{4.5}$ \\
 &
  T5-RG &
  $7.8_{0.5}$/$9.6_{1.0}$/$10.9_{0.7}$ &
  $21.3_{1.3}$/$26.1_{1.5}$/$29.6_{2.0}$ &
  $23.4_{0.7}$/$27.0_{1.0}$/$29.5_{1.2}$ &
  $47.1_{2.8}$/$57.8_{4.2}$/$63.5_{3.1}$ \\ \cmidrule(l){2-6} 
 &
  T5-KPT &
  $10.7_{2.4}$/$11.7_{0.9}$/$11.8_{0.9}$ &
  $32.4_{4.2}$/$35.1_{1.7}$/$34.9_{1.2}$ &
  $29.7_{2.6}$/$31.9_{0.8}$/$32.8_{0.6}$ &
  $67.6_{7.2}$/$72.2_{2.2}$/$71.2_{1.8}$ \\
 &
  T5-RG-KPT &
  $10.5_{1.7}$/$11.4_{0.4}$/$11.4_{1.0}$ &
  $32.7_{2.5}$/$35.2_{0.9}$/$34.6_{1.0}$ &
  $29.6_{1.5}$/$31.8_{1.0}$/$32.6_{0.7}$ &
  $69.4_{4.2}$/$72.9_{1.1}$/$70.9_{2.6}$ \\
 &
  T5-RG-$K_g$ &
  $11.5_{1.4}$/$11.4_{0.6}$/$12.2_{0.4}$ &
  $32.7_{2.7}$/$33.9_{1.0}$/$35.0_{0.9}$ &
  $29.6_{1.4}$/$31.0_{0.9}$/$32.5_{0.3}$ &
  $68.4_{5.1}$/$70.5_{1.7}$/$71.1_{2.1}$ \\
 &
  T5-RG-$K_n$ &
  $11.4_{0.5}$/$11.3_{0.9}$/$11.1_{0.9}$ &
  $33.8_{0.7}$/$33.7_{1.5}$/$33.8_{1.2}$ &
  $30.7_{0.5}$/$31.2_{1.2}$/$32.0_{0.6}$ &
  $70.1_{0.8}$/$70.0_{2.5}$/$69.1_{2.7}$ \\ \cmidrule(l){2-6} 
 &
  T5-RG-KPT(DGPT) &
  $10.0_{1.3}$/$11.4_{1.1}$/$11.4_{0.7}$ &
  $29.6_{3.0}$/$32.8_{2.9}$/$33.4_{1.2}$ &
  $27.5_{1.5}$/$30.5_{1.9}$/$31.7_{0.7}$ &
  $64.6_{5.9}$/$69.4_{3.0}$/$69.6_{2.5}$ \\
 &
  T5-RG-KPT(Rand) &
  $11.4_{1.3}$/$12.3_{0.4}$/$11.3_{1.1}$ &
  $32.6_{2.7}$/$34.6_{0.5}$/$33.8_{1.4}$ &
  $29.9_{1.3}$/$31.8_{0.2}$/$32.1_{1.0}$ &
  $68.4_{4.8}$/$72.2_{0.9}$/$69.4_{3.2}$ \\ \cmidrule(l){2-6} 
 &
  T5-RG-KPT($\alpha$=0.1) &
  $8.9_{1.9}$/$10.5_{0.5}$/$11.5_{0.9}$ &
  $29.1_{3.5}$/$32.1_{1.4}$/$33.0_{1.2}$ &
  $27.6_{2.1}$/$30.0_{1.1}$/$31.0_{1.1}$ &
  $63.5_{7.8}$/$69.4_{1.5}$/$71.1_{1.9}$ \\
 &
  T5-RG-KPT($\alpha$=0.2) &
  $10.9_{1.5}$/$11.2_{1.2}$/$11.2_{1.5}$ &
  $32.9_{2.9}$/$33.5_{2.2}$/$33.7_{1.5}$ &
  $29.7_{1.7}$/$30.9_{1.5}$/$31.9_{0.9}$ &
  $70.4_{5.4}$/$71.0_{3.3}$/$70.4_{3.5}$ \\
 &
  T5-RG-KPT($\alpha$=0.4) &
  $9.3_{0.7}$/$12.0_{0.4}$/$12.1_{1.1}$ &
  $29.4_{0.6}$/$34.9_{1.3}$/$35.3_{1.5}$ &
  $28.1_{0.5}$/$32.0_{0.7}$/$33.0_{0.8}$ &
  $61.9_{1.6}$/$71.5_{1.2}$/$71.8_{2.2}$ \\
 &
  T5-RG-KPT($\alpha$=1.0) &
  $10.4_{1.4}$/$11.9_{0.6}$/$12.1_{0.9}$ &
  $30.3_{3.7}$/$34.0_{1.5}$/$34.6_{1.0}$ &
  $28.4_{1.6}$/$31.3_{1.2}$/$32.6_{0.4}$ &
  $62.5_{6.9}$/$69.6_{2.4}$/$70.1_{2.2}$ \\ \midrule
\multirow{12}{*}{PC} &
  T5 &
  $1.8_{0.5}$/$2.2_{0.1}$/$2.5_{0.3}$ &
  $4.0_{0.7}$/$4.5_{1.2}$/$5.6_{1.0}$ &
  $14.7_{1.0}$/$15.4_{0.9}$/$16.3_{0.6}$ &
  $15.1_{4.2}$/$17.1_{1.3}$/$17.4_{3.5}$ \\
 &
  T5-RG &
  $2.8_{0.2}$/$2.9_{0.4}$/$3.4_{0.2}$ &
  $7.2_{0.7}$/$7.3_{0.3}$/$7.5_{0.4}$ &
  $17.4_{0.3}$/$18.0_{0.4}$/$17.8_{0.8}$ &
  $13.9_{4.2}$/$14.6_{3.3}$/$16.5_{1.6}$ \\ \cmidrule(l){2-6} 
 &
  T5-KPT &
  $2.9_{0.2}$/$3.1_{0.4}$/$3.3_{0.1}$ &
  $7.9_{0.7}$/$7.8_{0.7}$/$8.0_{0.4}$ &
  $17.5_{0.8}$/$17.9_{0.4}$/$18.2_{0.3}$ &
  $22.2_{4.5}$/$19.8_{3.9}$/$18.2_{4.1}$ \\
 &
  T5-RG-KPT &
  $2.9_{0.2}$/$3.2_{0.3}$/$3.4_{0.2}$ &
  $8.1_{0.7}$/$7.9_{0.7}$/$8.2_{0.6}$ &
  $17.9_{0.6}$/$18.2_{0.5}$/$18.3_{0.7}$ &
  $18.1_{6.6}$/$20.0_{3.0}$/$19.1_{2.7}$ \\
 &
  T5-RG-$K_g$ &
  $3.0_{0.4}$/$3.0_{0.3}$/$3.3_{0.2}$ &
  $7.7_{0.9}$/$7.4_{0.7}$/$7.5_{0.5}$ &
  $17.5_{0.6}$/$18.2_{0.2}$/$18.2_{0.6}$ &
  $21.5_{4.2}$/$14.4_{3.4}$/$17.9_{5.3}$ \\
 &
  T5-RG-$K_n$ &
  $2.8_{0.2}$/$3.1_{0.3}$/$3.4_{0.2}$ &
  $7.9_{0.9}$/$7.9_{1.1}$/$8.2_{0.8}$ &
  $17.8_{0.7}$/$18.0_{0.6}$/$18.1_{0.8}$ &
  $15.7_{4.9}$/$18.1_{2.9}$/$19.0_{4.3}$ \\ \cmidrule(l){2-6} 
 &
  T5-RG-KPT(DGPT) &
  $2.7_{0.2}$/$2.8_{0.2}$/$3.2_{0.3}$ &
  $7.8_{0.7}$/$7.6_{0.9}$/$8.0_{0.5}$ &
  $18.0_{0.3}$/$17.9_{0.5}$/$18.6_{0.4}$ &
  $14.1_{2.5}$/$18.3_{3.2}$/$17.0_{3.0}$ \\
 &
  T5-RG-KPT(Rand) &
  $2.9_{0.3}$/$3.0_{0.4}$/$3.4_{0.1}$ &
  $7.7_{0.4}$/$7.7_{1.0}$/$8.0_{0.7}$ &
  $18.2_{0.2}$/$17.9_{0.5}$/$18.4_{0.8}$ &
  $13.5_{3.7}$/$19.4_{2.5}$/$18.0_{1.4}$ \\ \cmidrule(l){2-6} 
 &
  T5-RG-KPT($\alpha$=0.1) &
  $2.9_{0.1}$/$3.2_{0.3}$/$3.6_{0.2}$ &
  $7.9_{0.8}$/$7.8_{0.6}$/$8.0_{0.4}$ &
  $18.0_{0.6}$/$18.1_{0.4}$/$18.4_{0.5}$ &
  $13.8_{4.9}$/$19.0_{2.9}$/$17.4_{2.8}$ \\
 &
  T5-RG-KPT($\alpha$=0.2) &
  $2.6_{0.4}$/$3.0_{0.3}$/$3.5_{0.1}$ &
  $7.7_{0.5}$/$7.8_{0.9}$/$8.3_{0.6}$ &
  $18.0_{0.3}$/$17.8_{0.5}$/$18.4_{0.6}$ &
  $13.8_{4.3}$/$19.7_{1.4}$/$19.1_{4.3}$ \\
 &
  T5-RG-KPT($\alpha$=0.4) &
  $2.9_{0.3}$/$2.9_{0.4}$/$3.5_{0.1}$ &
  $8.3_{0.7}$/$7.9_{0.6}$/$8.0_{0.6}$ &
  $17.9_{0.6}$/$17.8_{0.4}$/$18.5_{0.6}$ &
  $17.5_{5.3}$/$20.7_{3.2}$/$17.1_{3.2}$ \\
 &
  T5-RG-KPT($\alpha$=1.0) &
  $2.7_{0.2}$/$2.9_{0.4}$/$3.3_{0.1}$ &
  $7.8_{0.8}$/$7.7_{0.7}$/$8.2_{0.5}$ &
  $18.0_{0.3}$/$17.9_{0.5}$/$18.4_{0.8}$ &
  $15.4_{5.1}$/$19.8_{2.9}$/$19.1_{4.6}$ \\ \midrule
\multirow{12}{*}{WoW} &
  T5 &
  $3.3_{1.0}$/$2.9_{0.3}$/$3.0_{0.4}$ &
  $10.1_{1.5}$/$9.0_{0.4}$/$8.6_{0.9}$ &
  $14.8_{0.9}$/$14.4_{0.6}$/$14.1_{0.8}$ &
  $13.7_{4.7}$/$11.0_{1.1}$/$10.5_{1.7}$ \\
 &
  T5-RG &
  $2.6_{0.3}$/$3.0_{0.4}$/$3.5_{0.2}$ &
  $9.3_{0.3}$/$10.0_{0.8}$/$10.6_{0.3}$ &
  $15.1_{0.1}$/$15.7_{0.5}$/$16.3_{0.3}$ &
  $7.1_{1.0}$/$8.4_{1.4}$/$9.6_{0.8}$ \\ \cmidrule(l){2-6} 
 &
  T5-KPT &
  $4.7_{0.8}$/$3.5_{0.6}$/$4.3_{0.6}$ &
  $13.2_{0.7}$/$11.7_{0.7}$/$12.2_{0.9}$ &
  $17.5_{0.5}$/$16.7_{0.5}$/$17.4_{0.6}$ &
  $15.1_{2.9}$/$10.0_{1.6}$/$12.1_{2.2}$ \\
 &
  T5-RG-KPT &
  $4.9_{0.8}$/$4.1_{0.4}$/$4.7_{0.5}$ &
  $13.3_{0.7}$/$12.2_{0.7}$/$13.1_{0.7}$ &
  $17.6_{0.7}$/$17.2_{0.6}$/$18.0_{0.5}$ &
  $15.2_{3.1}$/$11.8_{1.8}$/$13.1_{1.7}$ \\
 &
  T5-RG-$K_g$ &
  $4.3_{0.2}$/$3.3_{0.5}$/$3.8_{0.5}$ &
  $12.3_{0.4}$/$11.0_{0.9}$/$12.1_{1.0}$ &
  $17.2_{0.2}$/$16.6_{0.5}$/$17.4_{0.6}$ &
  $14.7_{1.5}$/$10.1_{1.8}$/$11.4_{1.5}$ \\
 &
  T5-RG-$K_n$ &
  $4.2_{0.5}$/$3.6_{0.6}$/$4.1_{0.4}$ &
  $12.5_{0.8}$/$11.6_{1.0}$/$12.3_{0.4}$ &
  $17.1_{0.5}$/$16.6_{0.8}$/$17.4_{0.4}$ &
  $12.9_{2.6}$/$10.3_{2.1}$/$11.4_{1.3}$ \\ \cmidrule(l){2-6} 
 &
  T5-RG-KPT(DGPT) &
  $4.0_{1.0}$/$4.1_{0.5}$/$4.6_{0.3}$ &
  $11.8_{1.6}$/$12.1_{0.6}$/$12.7_{0.8}$ &
  $16.6_{1.2}$/$17.1_{0.4}$/$17.8_{0.5}$ &
  $12.2_{4.1}$/$11.9_{1.8}$/$13.0_{0.8}$ \\
 &
  T5-RG-KPT(Rand) &
  $4.1_{1.1}$/$3.7_{0.3}$/$4.0_{0.3}$ &
  $12.1_{1.6}$/$11.6_{0.3}$/$11.9_{0.3}$ &
  $16.9_{1.2}$/$16.9_{0.3}$/$17.3_{0.2}$ &
  $12.2_{3.6}$/$10.6_{0.5}$/$11.0_{1.2}$ \\ \cmidrule(l){2-6} 
 &
  T5-RG-KPT($\alpha$=0.1) &
  $3.7_{0.5}$/$3.6_{0.4}$/$3.8_{0.5}$ &
  $11.0_{1.0}$/$11.3_{0.8}$/$11.6_{0.4}$ &
  $16.1_{0.6}$/$16.5_{0.5}$/$16.9_{0.4}$ &
  $10.9_{2.0}$/$10.1_{1.2}$/$10.2_{1.5}$ \\
 &
  T5-RG-KPT($\alpha$=0.2) &
  $3.8_{1.0}$/$3.5_{0.4}$/$4.3_{0.3}$ &
  $12.0_{1.3}$/$11.4_{0.7}$/$12.4_{0.7}$ &
  $16.8_{1.0}$/$16.6_{0.6}$/$17.6_{0.3}$ &
  $11.7_{3.2}$/$10.0_{1.5}$/$11.8_{0.8}$ \\
 &
  T5-RG-KPT($\alpha$=0.4) &
  $4.2_{0.7}$/$3.9_{0.4}$/$4.3_{0.3}$ &
  $12.5_{0.9}$/$11.9_{0.7}$/$12.5_{0.8}$ &
  $17.1_{0.6}$/$16.9_{0.5}$/$17.6_{0.6}$ &
  $12.7_{2.6}$/$11.2_{1.2}$/$12.4_{1.0}$ \\
 &
  T5-RG-KPT($\alpha$=1.0) &
  $3.7_{1.0}$/$3.6_{0.4}$/$4.2_{0.2}$ &
  $11.5_{1.2}$/$11.3_{0.7}$/$12.3_{0.3}$ &
  $16.5_{0.8}$/$16.5_{0.5}$/$17.3_{0.3}$ &
  $11.9_{3.8}$/$11.4_{1.5}$/$12.1_{1.0}$ \\ \bottomrule
\end{tabular}
\caption{Performance of models based on T5-small. 
Results of 50/100/200-shot are separated by a slash. 
We report the means and standard deviations of 5 runs. ``DGPT" means using DialoGPT to extract keywords and ``Rand" refers to random keyword extraction. $\alpha$ is the keyword ratio (default 0.3).}
\label{tab:t5small_all}
\end{table*}

\end{document}